\documentclass[times, review, 10pt]{elsarticle}

\usepackage{textcomp}
\usepackage{stfloats}
\usepackage{verbatim}

\usepackage{tabularx,booktabs}
\usepackage{color}
\usepackage{tabularray}
\UseTblrLibrary{diagbox}

\usepackage{tikz}
\usetikzlibrary{patterns,backgrounds}

\usepackage{amsmath}
\usepackage{amsfonts}
\usepackage{amssymb}
\usepackage{amsthm}
\usepackage{mathrsfs}
\usepackage{extarrows}
\usepackage{cancel}

\usepackage{url}

\usepackage{multirow} 

\usepackage[colorlinks,linkcolor=blue,anchorcolor=blue,citecolor=blue]{hyperref}

\newtheorem{example}{Example}

\usepackage[linesnumbered,ruled,lined]{algorithm2e}

\usepackage{subcaption}

\usepackage{float} 

\usepackage{booktabs}

\begin{document}

\begin{frontmatter}

\title{Evaluating Evidential Reliability In Pattern Recognition Based On Intuitionistic Fuzzy Sets}

\author[uestc,uestc2]{Juntao Xu}
\author[uestc2]{Tianxiang Zhan}
\author[uestc2,vu]{Yong Deng \corref{cor}}

\affiliation[uestc]{
    institute = {Glasgow College},
    organization={University of Electronic Science and Technology of China},
    city={Chengdu},
    postcode={610054},
    state={Sichuan},
    country={China}}

\affiliation[uestc2]{
    institute = {Institute of Fundamental and Frontier Science},
    organization={University of Electronic Science and Technology of China},
    city={Chengdu},
    postcode={610054},
    state={Sichuan},
    country={China}}

\affiliation[vu]{
    institute = {School of Medicine},
    organization={Vanderbilt University},
    city={Nashville},
    postcode={37240},
    state={TN},
    country={USA}}

\cortext[cor]{Corresponding author: Yong Deng (e-mail: dengentropy@uestc.edu.cn)}

\begin{abstract}
Determining the reliability of evidence sources is a crucial topic in Dempster-Shafer theory (DST). Previous approaches have addressed high conflicts between evidence sources using discounting methods, but these methods may not ensure the high efficiency of classification models. In this paper, we consider the combination of DS theory and Intuitionistic Fuzzy Sets (IFS) and propose an algorithm for quantifying the reliability of evidence sources, called Fuzzy Reliability Index (FRI). The FRI algorithm is based on decision quantification rules derived from IFS, defining the contribution of different BPAs to correct decisions and deriving the evidential reliability from these contributions. The proposed method effectively enhances the rationality of reliability estimation for evidence sources, making it particularly suitable for classification decision problems in complex scenarios. Subsequent comparisons with DST-based algorithms and classical machine learning algorithms demonstrate the superiority and generalizability of the FRI algorithm. The FRI algorithm provides a new perspective for future decision probability conversion and reliability analysis of evidence sources.
\end{abstract}


\end{frontmatter}

\section{Introduction}
Evidence theory, also referred to as Dempster-Shafer theory (DST) \cite{Dempster, Shafer}, has been extensively applied across multiple fields, including decision-making \cite{decisionmaking3,decisionmaking4}, fault diagnosis \cite{faultdiognosis1,faultdiagnosis2}, clustering \cite{zhang2022bsc, Liuzhe2024clustering} and pattern recognition \cite{zhang2024divergence,deng2024random}. The ability of DST to effectively manage uncertain, imprecise, and conflicting information \cite{conflict1,conflict2} has made it a favored tool among both researchers and practitioners. Consequently, it has been further developed into various areas, such as complex evidence theory \cite{Xiao2023NQMF,Xiao2023Acomplexweighted} and generalized quantum evidence theory \cite{Xiao2023QuantumXentropy}. In recent studies, advancements in evidence theory have been linked with the exploration of information fractal dimensions to assess the complexity of mass functions \cite{Qiang2022fractal,zhan2024generalized}, as well as new entropy measures such as Deng entropy \cite{zhao2024linearity}, generalized information entropy \cite{zhan2024generalized}, Tsallis extropy \cite{balakrishnan2022tsallis} and Deng extropy \cite{DengeXtropy}.

However, despite its advantages, DST has several notable limitations. One of the primary issues is its tendency to produce counter-intuitive results \cite{counterintuitive}, particularly when dealing with highly conflicting evidence \cite{highlyconflict1, highlyconflict3}, as highlighted by the Zadeh paradox \cite{zadehparabox}. This paradox illustrates how the combination of seemingly reliable evidence can lead to absurd conclusions, challenging the robustness and reliability of DST in practical applications. The robustness of DST is another concern, as it often struggles to maintain consistent performance in the face of varying data quality and conflicting sources. Researchers have identified that the reliability of the body of evidence (BoE) can significantly impact the outcomes of DST-based analyses \cite{Reliability1,Reliability2}. As a result, numerous studies have focused on developing methods to improve the robustness and reliability of DST by addressing the quality and reliability of evidence sources . Historically, scholars have approached these challenges from various angles. Early works by Dempster and Shafer laid the foundation for DST, providing a formal framework for combining evidence from different sources \cite{Dempster,Shafer}. Later, researchers like Yager and Smets proposed modifications to the combination rules to handle conflicting evidence better \cite{Yager, Smets}. Yager introduced an alternative combination rule that accommodates conflict by redistributing it among the evidence, while Smets developed an unnormalized combination rule to preserve certain desirable properties. Further contributions were made by Murphy, who suggested averaging the evidence to mitigate conflict \cite{Murphy}, and Deng et al., who introduced a distance-based weighted average approach \cite{Deng}. 
Xiao et al. presented belief divergence to modify evidences to improve the performance of information fusion \cite{Xiao2023Generalizeddivergence,huang2023higherR}. Liu et al. introduced an approach to assess dissimilarity \cite{Liu}. Jiang developed an approach for calculating the correlation coefficient of belief functions \cite{Jiang}.

In parallel, Fuzzy Sets (FS) have been extensively utilized to address uncertainty and imprecision by introducing a degree of membership, which allows for a more flexible representation of data that does not strictly belong to one set or another. This capability has found applications in various fields. For example, Fuzzy Neural Networks (FNN) combine the ability of neural networks to learn from data with the capacity of fuzzy logic to handle uncertainty \cite{Bronik2024fuzzyneuralnetwork}. Additionally, fuzzy clustering has been widely applied in areas such as decision making \cite{2022mahmoodfuzzysetdecisionmaking} and image segmentation \cite{Wu2023fuzzyclustering}. Based on that, Intuitionistic Fuzzy Sets (IFS) extends traditional fuzzy set theory by incorporating the degree of non-membership, providing a more nuanced characterization of uncertainty \cite{IFS1,IFS2}.

Inspired by IFS, this paper proposes an evidence source reliability measurement algorithm called Fuzzy Reliability Index (FRI). Specifically, the use of FRI for evidence fusion involves five steps: first, using Triangular Fuzzy Numbers (TFNs) to convert the numerical values of sample features into corresponding BPAs; then, applying the transformation rules established between DS theory and IFS to translate BPAs into corresponding Intuitionistic Fuzzy Values (IFVs); based on this, the quantification scale of IFS for decision-making is used to determine the contribution of BPAs to correct decisions for evaluating the reliability of the corresponding evidence sources; subsequently, the relative reliability of evidence sources is converted into final discount factors for evidence fusion through a normalization process, and BPAs are adjusted accordingly in subsequent decisions; finally, BPAs are fused using DST. In the experimental phase, we utilize numerical examples and pattern classification applications to evaluate FRI against other DST-based algorithms as well as conventional machine learning techniques, illustrating the proposed method's enhanced performance and broad applicability. The key contributions of this study are outlined as follows:

1. Defining the transformation rules between IFS and DST, thus more reasonably and accurately quantifying the contribution of BPAs to correct decisions.

2. Defining a new metric based on the decision scale of IFS, which provides a new perspective for the probability transformation of BPAs.

3. The proposed reliability evaluation method enhances the importance of normal BPAs in decision-making while effectively reducing the adverse impact of abnormal BPAs, making it particularly suitable for pattern classification problems in complex scenarios.

The structure of this paper is as follows. In Section \hyperref[section 2]{2}, the relevant fundamentals of DST, IFS, and TFN will be introduced. In Section \hyperref[section 3]{3}, the FRI algorithm will be proposed and illustrated in detail with several numerical examples. Section \hyperref[section 4]{4} will focus on demonstrating the effectiveness of the FRI algorithm through various applications in pattern recognition and comparing it with other algorithms, followed by some discussion. The final \hyperref[conclusion]{section} provides some conclusions and future work.

\section{Preliminaries}
\label{section 2}
\subsection{Dempster-Shafer evidence theory (DST)}
The challenge of modeling the uncertainty and dynamics of systems \cite{Wang2020CommunicatingSA, wang2022modelling} has garnered significant attention. Dempster-Shafer evidence theory (DST), also referred to as the theory of belief functions, presents an intriguing alternative to conventional probability theory for managing uncertainty. It allows us to assign belief to sets of possibilities, making it incredibly useful when dealing with incomplete or ambiguous information. The following sections will provide an overview of DST.
\subsubsection{Basic Conceptions}
\noindent  \textbf{(1) Frame of Discernment (FOD)} \cite{Dempster}
\par
The frame of discernment (FOD) $\Theta$ is one of the fundamental concepts of DST. Here, $\Theta$ represents the exhaustive set of all possible values of a random variable.

\par
\begin{equation}
\Theta = \{ A_1, A_2, \ldots, A_n \}
\label{eq:frame_of_discernment}
\end{equation}

According to the properties of sets, all elements of $\Theta$ are exhaustive and mutually exclusive. $2^\Theta$ denotes the power set of $\Theta$, where
\par
\begin{equation}
2^\Theta=\left\{\varnothing,\left\{A_1\right\},\left\{A_2\right\} \ldots\left\{A_n\right\},\left\{A_1, A_2 \ldots A_i\right\} \ldots \Theta\right\}
\label{eq:power_set}
\end{equation}

In this power set, each element of a set corresponds to a possible scenario regarding the values taken by the variable.
\\ \par
\noindent  \textbf{(2) Basic Probability Assignment (BPA)} \cite{Dempster}
\par
For each element in $2^\Theta$, the basic probability assignment (BPA) function is a mapping $2^\theta \rightarrow[0,1]$, which satisfies the following two conditions:

\par
\begin{equation}
m(\emptyset)=0
\label{eq:bpa_empty_set}
\end{equation}

\par
\begin{equation}
\sum_{A \in \Theta} m(A)=1
\label{eq:bpa_sum}
\end{equation}

\par
Let \(m\) be the basic probability assignment (BPA) function on \(A\). For any subset \(A\) of \(\Theta\), \(m(A)\) represents the basic probability mass associated with \(A\). When \(A \subseteq \Theta\) and \(A \neq \Theta\), \(m(A)\) reflects the degree of belief in the proposition \(A\). When \(A = \Theta\), \(m(A)\) signifies the uncertainty regarding the distribution of the probability mass. In the belief framework, since the empty set contains no elements, it is not assigned any support, which implies \(m(\emptyset) = 0\).
\\ \par
\noindent  \textbf{(3) Belief ($Bel$) and Plausibility ($Pl$) Function} \cite{Shafer}
\par Given a FOD \(\Theta\), the belief function $(Bel)$ and the plausibility ($Pl$) function can be defined as follows:

\par 
\begin{equation}
Bel(A)=\sum_{B \in A} m(B) \quad \forall A \subseteq \Theta
\end{equation}

\par 
\begin{equation}
Pl(A)=\sum_{B \cap A \neq \emptyset} m(B) \quad \forall A \subseteq \Theta
\end{equation}

The difference between Bel and Pl is that the belief function represents the total belief committed to \(A\), whereas the plausibility function represents the degree to which \(A\) is not disbelieved. Their relationship can be expressed as follows:

\par 
\begin{equation}
Pl(A)=1-Bel(\bar{A})
\end{equation}

\par 
\begin{equation}
P l(A) \geq Bel(A)
\end{equation}

where \(\bar{A}\) represents the complement of \(A\) in FOD.
Because of the meanings and relationships between the $Bel$ and the $Pl$, they can be used as the lower and upper bounds of the proposition \(A\), denoted as \(A[\text{$Bel$}(A), \text{$Pl$}(A)]\).

\subsubsection{Dempster's Combination Rules}
\par For two BPAs \( m_1 \) and \( m_2 \) defined on FOD, Dempster's Combination Rules \cite{Dempster} are defined as:

\par 
\begin{equation}
\left(m_1 \oplus m_2\right)(A)=\frac{1}{1-K} \sum_{B, C \subseteq \Theta, B \cap C=A} m_1(B) m_2(C),
\end{equation}

where $K$ represents the conflict between $m_1$ and $m_2$:

\par 
\begin{equation}
K=\sum_{B, C \subseteq \Theta, B \cap C=\emptyset} m_1(B) m_2(C)
\end{equation}




\subsubsection{Pignistic Probability Transfomation (PPT)}
Pignistic Probability Transformation (PPT) \cite{PPT1} is a method used to apply the BPA framework in decision-making contexts. Its principle involves averaging the uncertainty associated with each proposition across all its elements, assuming no additional external information is available. This ensures fairness in decision-making. Specifically, it can be expressed as:

\par 
\begin{equation}
\operatorname{BetP}(A)=\sum_{B \subseteq \Theta, A \in B} \frac{1}{|B|} \cdot \frac{m(B)}{1-m(\emptyset)},
\end{equation}

where $|B|$ is the number of elements in $B$.

\subsubsection{Discounting Rules}

Because the reliability of different sources of evidence varies, it is necessary in the decision-making process to discount them to different extents based on their reliability \cite{Discounting1}. If the reliability of a source of evidence is denoted as \( \lambda \in [0, 1] \), one of the classic discounting operations proposed by Shafer \cite{Shafer} is defined as:

\par 
\begin{equation}
\left\{\begin{array}{l}
m^{\prime}(A)=\lambda m(A), A \subseteq \Theta \\
m^{\prime}(\Theta)=1-\lambda+\lambda m(\Theta)
\end{array}\right.
\end{equation}

where \( \lambda \) represents the evidential reliavility. As the reliability of an evidence decreases, the mass allocated to \( \Theta \) increases. When the reliability of this evidence is 0, the BPA becomes \( m(\Theta) = 0 \), which means all degrees of belief are completely uncertain.

\subsection{Atanassov’s Intuitionistic Fuzzy Sets (IFS)}
Atanassov's Intuitionistic Fuzzy Sets (IFS), introduced by K. Atanassov \cite{IFS1,IFS2} in 1986. While fuzzy sets only account for the degree to which an element belongs to a set, IFS additionally quantify the degree of non-belonging and the uncertainty or hesitation regarding the membership. This enhanced framework allows IFS to better handle and represent uncertainty.
\subsubsection{Basic conceptions}

\noindent  \textbf{(1) Membership, Non-membership and Hesitation Degree} \cite{IFS1, IFS2}
\par
Let \( X = \{x_1, x_2, \ldots, x_n\} \) be a universe of discourse. According to the definition, each \( A \) in \( X \) can be expressed as:
\par 
\begin{equation}
A=\left\{\left\langle x, \mu_A(x), v_A(x)\right\rangle \mid x \in X\right\}
\end{equation}
\par where $\mu_A(x): X \rightarrow[0,1]$ and $v_A(x): X \rightarrow[0,1]$ represent the membership and non-membership degrees of \( x \) to \( A \), respectively. The relationship between \( \mu_A(x) \) and \( v_A(x) \) satisfies the following:

\par 
\begin{equation}
0 \leq \mu_A(x)+v_A(x) \leq 1
\end{equation}

$\pi_A(x)$ represents the  hesitancy degree, used when the membership relationship between $x$ and $A$ cannot be determined, and is expressed as:

\par 
\begin{equation}
\pi_A(x)=1-\mu_A(x)-v_A(x)
\end{equation}
\par Here, the value range of $\pi_A(x)$ is $[0,1]$. A higher value of $\pi_A(x)$ indicates greater ambiguity in the judgement of $x$.
\par It is worth noting that there are other methods to represent intuitionistic fuzzy sets. For instance, based on the relationship between \( \mu_A(x) \) and \( v_A(x) \) in equation (15), it can be derived that \( \mu_A(x) \leq 1 - v_A(x) \). Therefore, another common representation for \( A \) in IFS is in the form of an interval \( [\mu_A(x), 1 - v_A(x) ]\) \cite{IFSrepresentation1}.
\\ \par
\noindent  \textbf{(2) Intuitionistic Fuzzy Values (IFV)} \cite{IFV}
\par In IFS, the membership relationship between \( A \) and \( X \) can be represented using \( \mu_X \), \( v_X \), and \( \pi_X \). When \( |X| = 1 \), indicating that there is only one element contained in \( X \), the set reduces to a singleton, transforming the intuitionistic fuzzy set into an intuitionistic fuzzy values (IFV), denoted as \( A = \left\langle \mu_A, v_A \right\rangle \).
\par For example, when \( \left\langle \mu_A(x), v_A(x) \right\rangle = \langle 0.1, 0.2 \rangle \), this can be understood as the possibility or support degree of \( x \) belonging to \( A \) being 0.1, and the possibility of \( x \) not belonging to \( A \) being 0.2. In this case, \( \pi(x) \) is 0.7, indicating that there is 0.7 uncertainty in determining whether \( x \) belongs to \( A \) or not.

\subsubsection{Ranking method of IFVs}
In decision-making problems, the ranking of IFVs is crucial because it determines how to make the most rational judgments when variables have multiple scenarios and IFVs associated with them.

\noindent  \textbf{(1) Normal Ranking} \cite{IFS2}
\par Suppose there are two intuitionistic fuzzy values  $A=\left\langle\mu_A, v_A\right\rangle$ and $B=\left\langle\mu_B, v_B\right\rangle$. It can be easily determined the priority order between A and B in the following scenarios.
\par 
\begin{equation}
\left\{\begin{array}{l}\mu_A(x) \geq \mu_B(x) \\ v_A(x) \leq v_B(x)\end{array} \Leftrightarrow A \geq B\right.
\end{equation}
\par However, when the situation becomes more complex, the ranking order between A and B may not be so straightforward. Therefore, below score functions and accuracy functions are defined to properly compare the rankings of IFVs.
\\ \par
\noindent  \textbf{(2) Score Function and Accuracy Function} \cite{IFVranking1}
\par Given a IFV \( A = \left\langle \mu_A, v_A \right\rangle \). Then the score function $S$ and accuracy function $H$ of \( A \) are defined as:

\par 
\begin{equation}
S_A(x)=\mu_A(x)-v_A(x)
\end{equation}

\par 
\begin{equation}
H_A(x)=\mu_A(x)+v_A(x)=1-\pi_A(x)
\end{equation}

\par The score function \( S_A(x) \) can be understood as the degree of support for determining \( x \) belongs to the IFS-A. The larger \( S_A(x) \), the greater the support for \( x \) belonging to \( A \).
\par The accuracy function reflects the degree of certainty in making this determination, which is related to the hesitation degree (\( \pi_A(x) \)) in IFS. The larger \( H_A(x) \), the greater the certainty that \( x \) belongs to IFS-A. When the hesitation degree is 1, \( H_A(x) \) becomes 0, indicating that the determination is entirely baseless.
\par Based on the mentioned score function and accuracy function, when faced with the problem of ranking multiple IFVs, the following rules can be applied.
\par
Let $A=\left\langle\mu_A, v_A\right\rangle$ and $B=\left\langle\mu_B, v_B\right\rangle$ be two IFVs, and \( S_A(x) \), \( H_A(x) \), \( S_B(x) \), \( H_B(x) \) represent their score functions and accuracy functions respectively, in order to get their rankings in decision-making problem.\\
(1) if $S_A(x)>S_B(x)$, then \( x \) is more likely to belong to \( A \), denoted as \( A > B \).\\
(2) if $S_A(x)=S_B(x)$, then \\
(i) if $H_A(x)>H_B(x)$, \( x \) is more likely to belong to \( A \), denoted as \( A > B \).\\
(ii) if $H_A(x)<H_B(x)$, \( x \) is more likely to belong to \( B \), denoted as \( A < B \).\\
(iii) if $H_A(x)<H_B(x)$, \( A \) and \( B \) have the same likelihood, denoted as \( A = B \).




\subsection{Triangular Fuzzy Numbers (TFN)}
    Triangular Fuzzy Numbers \cite{TFN1} are an effective tool for representing and handling uncertainty. TFNs are represented by a triplet \((a, b, c)\), where \(a \leq b \leq c\) correspond to the lower bound, midpoint, and upper bound of the fuzzy number, respectively. The membership function $\mu_{\tilde{A}}(x)$ of a TFN is defined as follows:

\begin{equation}
\mu_{\tilde{A}}(x) =
\begin{cases}
0 & \text{if } x < a \\
\frac{x - a}{b - a} & \text{if } a \leq x \leq b \\
\frac{c - x}{c - b} & \text{if } b < x \leq c \\
0 & \text{if } x > c
\end{cases}
\end{equation}

\par where \(x\) represents the value of the attribute for the object.

\par Given a set of samples, for a specific attribute, the minimum value, maximum value, and average value can be determined. Based on these three attribute values, a triangular fuzzy number can be established to describe the proposition.


\section{Proposed Method: Fuzzy Reliability Index Based on IFS}
\label{section 3}
In the following method, we assess the evidential reliability based on the similarities between DST and IFS, which is called Fuzzy Reliability Index (FRI). By leveraging the strengths of both frameworks, we can estimate the reliability of evidence more effectively.

\subsection{BPAs generated from TFNs}
\par The generation of BPAs is achieved through triangular fuzzy numbers, following the strategy outlined below:

1. When a sample intersects the TFN model for a proposition, the y-coordinate of the intersection is recorded to represent the degree of membership supporting that proposition.

2. When a sample intersects multiple TFN models for different propositions, the highest y-coordinate is recorded to represent the degree of membership supporting that proposition, while the lowest y-coordinate is used to represent the membership degree for the proposition encompassing multiple subsets.

3. When a sample is within the TFN models of multiple propositions, the highest y-coordinate is recorded as the membership degree supporting each single proposition, and the lowest y-coordinate is the membership degree supporting the proposition of multiple subsets.

4. The BPA for each proposition is generated by normalizing the membership degrees obtained above, ensuring that the total sum of BPAs equals 1.

5. If a sample does not intersect with any triangular fuzzy number models of the propositions, it is considered that no valid information is obtained, and all BPA is assigned to the universal set, i.e., \( m(\Theta) = 1 \).

\begin{example}
Assume there is an object with three possible classes A, B, and C. Their triangular fuzzy numbers are \(A = (1,2,4)\), \(B = (0,1,3)\), and \(C = (1,3,5)\), respectively.

\begin{figure}[H]
    \centering
    \includegraphics[width=0.75\linewidth]{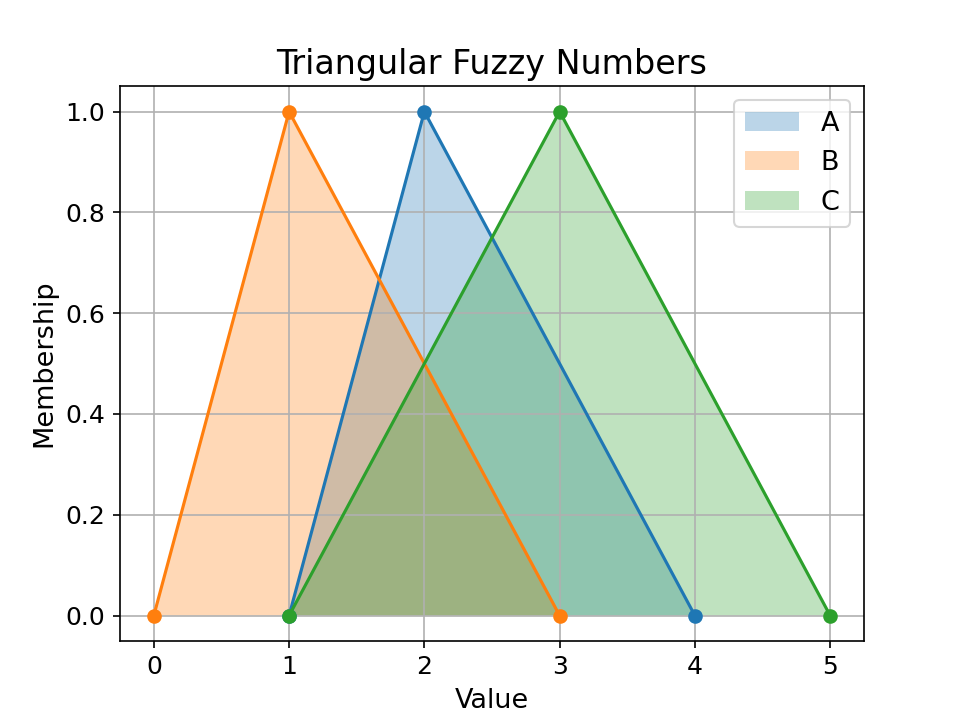}
    \caption{TFNs of different classes}
    \label{fig:enter-label}
\end{figure}

\par For a sample \(x\), the intersection points (i.e., membership degrees) with the three triangular fuzzy numbers are respectively expressed as \(\mu_{\tilde{A}}(x)\), \(\mu_{\tilde{B}}(x)\), and \(\mu_{\tilde{C}}(x)\).\\
When $x_1 = 0.5$,
\begin{equation}
\mu_{\tilde{A}}(x_1) = 0.5 \quad \mu_{\tilde{B}}(x_1) = 0 \quad \mu_{\tilde{C}}(x_1) = 0
\end{equation}
Then,
\begin{equation}
m(A)=\frac{0.5}{0.5+0+0} = 1 \quad m(B)=\frac{0}{0.5+0+0} = 0 \quad m(C)=\frac{0}{0.5+0+0} = 0
\end{equation}
When $x_2 = 1.5$,
\begin{equation}
\mu_{\tilde{A}}(x_2) = 0.5 \quad \mu_{\tilde{B}}(x_2) = 0.75 \quad \mu_{\tilde{C}}(x_2) = 0.25
\end{equation}
Then,
\begin{equation}
m(A)=\frac{0.5}{0.5+0.75+0.25} = 0.33 \quad m(B)=\frac{0.75}{0.5+0.75+0.25} = 0.5
\end{equation}
\begin{equation}
m(A,B,C)=\frac{0.25}{0.5+0.75+0.25} = 0.17
\end{equation}
When $x_3 = 2$,
\begin{equation}
\mu_{\tilde{A}}(x_3) = 1 \quad \mu_{\tilde{B}}(x_3) = 0.5 \quad \mu_{\tilde{C}}(x_3) = 0.5
\end{equation}
Then,
\begin{equation}
m(A)=\frac{1}{1+0.5+0.5} = 0.5 \quad \quad m(A,B,C)=\frac{0.5+0.5}{1+0.5+0.5} = 0.5
\end{equation}
When $x_4 = 3.5$,
\begin{equation}
\mu_{\tilde{A}}(x_4) = 0.25 \quad \mu_{\tilde{B}}(x_4) = 0 \quad \mu_{\tilde{C}}(x_4) = 0.75
\end{equation}
Then,
\begin{equation}
m(C)=\frac{0.75}{0.25+0+0.75} = 0.75 \quad \quad m(A,C)=\frac{0.25}{0.25+0+0.75} = 0.25
\end{equation}
\end{example}

\par \quad
\subsection{Similarities between DST and IFS}
\par In 2.1.1 (3), \( \text{$Bel$}(A) \) and \( \text{$Pl$}(A) \) respectively represent the degree to which \( A \) is supported and the degree to which \( A \) is not contradicted. In 2.2.1 (1), \( \mu_A(x) \) and \( v_A(x) \) indicate the degree of \( x \) belonging and not belonging to \( A \).


In decision-making problems, when assessing the relationship between \( x \) and \( A \), a higher degree of support for \( A \) implies that \( x \) is more likely to belong to \( A \). Conversely, if \( A \) is highly contradicted, it indicates a greater non-membership degree of \( x \) to \( A \). Therefore, due to the similarity in objectives, DST and IFS can be connected, leading to the following relationships:

\par 
\begin{equation}
Bel(A) = \mu_A(x)
\end{equation}

\par 
\begin{equation}
Pl(A) = 1 - v_A(x)
\end{equation}

\par Based on their relationships, the interval representation of \( A \) with its upper and lower bounds, \( A[\text{$Bel$}(A), \text{$Pl$}(A)] \), and the representation \( [\mu_A(x), 1 - v_A(x)] \) are essentially equivalent.

\subsection{Explanation of Representation Methods in Classification Problems}
\par For classification problems, it is essential to first clarify certain notations before introducing the methods.

\begin{itemize}
    \item $\Theta=\left\{A_1, A_2, A_3 \cdots A_k\right\}$ represents the frame of discernment (FOD), indicating the different types that a variable may take.
    \item \(O = \{O_1, O_2, \ldots, O_n\}\) represents the objects being evaluated.
    \item $S = \{ S_1, S_2, \ldots, S_j \}$ represents a set of evidence sources, which in classification problems could correspond to different features of an object.
    \item $m_j\left\{O_n\right\}\left(A_k\right)$ represents the support degree with which evidence source \( S_j \) judges object \( O_n \) to belong to \( A_k \), expressed in the form of BPA.
    \item $m^*\left\{O_n\right\}\left(A^*\right)$ represents the true class of object \( O_n \) as \( A^* \). Note that in supervised learning processes, the true class of objects is known. Therefore, during training, for $\forall A_k \in \Theta$ but \( A_k \neq A^* \), \( m^* = 0 \).
\end{itemize}

\subsection{Decision Confidence}
\par In 2.2.2 (2), the score function \( S \) is designated to resolve the ranking of different IFVs, and its calculation method is shown in Eq. 18. Similarly, based on the transformation relationships given by Eq.20 and 21, the decision confidence in DST, denoted as:

\par 
\begin{equation}
d c_j\left\{O_n\right\}\left(A_k\right)=\mu_{j k}-v_{j k}=\left(B e l+P l)_j\left\{O_n\right\}\left(A_k\right)-1\right.
\end{equation}

\par where $j$ indicates different sources and $k$ indicates different classes that object belongs to.

\subsection{Correct Decision Contribution}
For an evidence source \( S_j \), there may be different judgments regarding the classification of object \( O_n \) and the corresponding decision confidence. The contribution of this evidence source to the correct classification result is defined as follows:

\par 
\begin{equation}
C_{j n}=d c_j\left\{O_n\right\}\left(A^*\right)-\frac{\sum_{A_k \in\left\{\Theta-A^*\right\}} d c_j\left\{O_n\right\}(A_k)}{\mid \Theta-A^* \mid}
\end{equation}

\par where $\mid \Theta-A^* \mid$ denotes the number of elements in $\left\{\Theta-A^*\right\}$.
\par The correct decision contribution is the sum of positive and negative contributions. If an evidence source incorrectly supports an incorrect result, it increases its negative contribution to the correct decision. Since the final decision will only result in one outcome, to ensure that negative contributions are not overcounted, it is necessary to average the negative contributions across all incorrect classifications.

\subsection{Source Reliability}
An evidence source's reliability is defined by its contribution to the correct result. Evaluating each object's contribution \( \{O_1, O_2, \ldots, O_n\} \) allows for the calculation of the reliability \( R_j \) of the evidence source \( S_j \):

\par 
\begin{equation}
R_j=\sum_{i=1}^n C_{j i}
\end{equation}

\subsection{Source Weights}
Given the different reliability of evidence sources to correct decisions \( R_j \), their weights in subsequent decision stages are defined as follows:
\par 
\begin{equation}
W_j=\frac{R_j-\min \{R\}}{\max \{R\}-\min \{R\}}, \quad R \in\left\{R_1 , R_2 \cdots R_j\right\}
\end{equation}
\par Considering that evidence sources may have negative reliability towards correct decisions, meaning they strongly support a class other than the correct one, the calculation of evidence source weights is based on relative reliability to avoid negative weights. When an evidence source has higher reliability, it carries more weight in subsequent classification decisions. Conversely, when an evidence source has minimal reliability, its weight is zero, implying it is disregarded entirely in decision-making.

\begin{example}
Let $O_1$ be an object to be identified and \( S_1, S_2, S_3, S_4\) be three evidence sources in a FOD \( \Theta = \{ A, B, C\} \), and assume \( m^*\left\{ O_n \right\}\left( B \right) = 1 \).
\begin{equation}
m_1\{O_1\}(A)=0.4\quad m_1\{O_1\}(B)=0.2\quad m_1\left\{O_1\right\}(C)=0.2\quad m_1\{O_1\}(B, C)=0.2
\end{equation}
\begin{equation}
m_2\{O_1\}(A)=0.1\quad m_2\{O_1\}(B)=0.5\quad m_2\left\{O_1\right\}(C)=0.1\quad m_2\{O_1\}(A, C)=0.3
\end{equation}
\begin{equation}
 m_3\{O_1\}(A)=0.15\quad m_3\{O_1\}(B)=0.55\quad m_3\left\{O_1\right\}(A, C)=0.2\quad m_3\{O_1\}(\Theta)=0.1
\end{equation}
\begin{equation}
 m_4\{O_1\}(A)=0.3\quad m_4\{O_1\}(B)=0.3\quad m_4\left\{O_1\right\}(C)=0.3\quad m_4\{O_1\}(\Theta)=0.1
\end{equation}

\par First, compute the $Bel$ and $Pl$ functions for \( S \), representing them as $[Bel, Pl]$:


\begin{equation}
[\mathrm{Bel}, \mathrm{Pl}] =
\begin{matrix}
& A & B & C \\
\left. \begin{matrix} S_1 \\ S_2 \\ S_3 \\ S_4 \end{matrix} \right(
& \begin{matrix} [0.4, 0.4] \\ [0.1, 0.4] \\ [0.15, 0.45] \\ [0.3, 0.4] \end{matrix} 
& \begin{matrix} [0.2, 0.4] \\ [0.5, 0.5] \\ [0.55, 0.65] \\ [0.3, 0.4] \end{matrix}
& \left. \begin{matrix} [0.2, 0.4] \\ [0.1, 0.4] \\ [0, 0.3] \\ [0.3, 0.4] \end{matrix} \right)_{O_1}
\end{matrix}
\end{equation}


\par Based on the transformation relationship between DST and IFS, the decision confidence for different evidence sources making different classification decisions can be calculated as:


\begin{equation}
dc =
\begin{array}{cccc}
& A & B & C \\
\left. \begin{array}{c} S_1 \\ S_2 \\ S_3 \\ S_4 \end{array} \right(
& \begin{array}{c} -0.2 \\ -0.5 \\ -0.4 \\ -0.3 \end{array} 
& \begin{array}{c} -0.4 \\ 0 \\ 0.2 \\ -0.3 \end{array}
& \left. \begin{array}{c} -0.4 \\ -0.5 \\ -0.7 \\ -0.3 \end{array} \right)_{O_1}
\end{array}
\end{equation}

\par By comparing with the correct classification result, the classification contributions of the three evidence sources can be determined as:
\begin{equation}
 C_{11}=(-0.4)-\frac{(-0.2)+(-0.4)}{2}=-0.1
\end{equation}
\begin{equation}
 C_{21}=0-\frac{(-0.5)+(-0.5)}{2}=0.5
\end{equation}
\begin{equation}
 C_{31}=0.2-\frac{(-0.4)+(-0.7)}{2}=0.75
\end{equation}
\begin{equation}
 C_{41}=-0.3-\frac{(-0.3)+(-0.3)}{2}=0
\end{equation}

\par Since the calculation of contributions is closely related to the decision results, it is clear that \( S_2 \) and \( S_3 \) can make correct decisions, so their contribution values are positive. \( S_1 \) makes incorrect decisions, so it has a negative impact on the contribution value. \( S_4 \), being unable to determine the final result, is considered to have made no contribution to the decision. Furthermore, by comparing \( S_2 \) and \( S_3 \), since \( S_2 \) has a higher support for \( B \), the contribution of \( S_2 \) will be greater than that of \( S_3 \).
\par In terms of reliability analysis, since \( C_{31} > C_{21} > C_{41} > C_{11} \), \( S_3 \) is considered the most reliable, while \( S_1 \) is the least reliable.
\end{example}

\begin{example}
Let \( O_1, O_2, O_3\) be three objects to be identified and \( S_1, S_2, S_3\) be three evidence sources with their decision contribution \( C_1, C_2, C_3\) shown as follows:

\begin{equation}
C = 
\begin{array}{cccc}
 & O_1 & O_2 & O_3 \\
 \left. \begin{array}{c} C_1 \\ C_2 \\ C_3 \end{array} \right (
&  \begin{array}{c} 0.5 \\ 0.2 \\ 0.25  \end{array} 
& \begin{array}{c}  0.6 \\ -0.25 \\ 0.4  \end{array}
& \left. \begin{array}{c} 0.3 \\ -0.15 \\ 0.1 \end{array} \right) \\
\end{array} 
\end{equation}

\par Then,
\begin{equation}
C_1=0.5+0.6+0.3=1.4 \quad C_2=0.2+(-0.25)+(-0.15)=-0.2 \quad C_3=0.25+0.4+0.1=0.75    
\end{equation}

\par According to the relative weight calculation method:
\begin{equation}
W_1=\frac{1.4-(-0.2)}{1.4-(-0.2)}=1 \quad W_2=\frac{-0.2-(-0.2)}{1.4-(-0.2)}=0 \quad W_3=\frac{0.75-(-0.2)}{1.4-(-0.2)}=0.59375
\end{equation}

\end{example}
\par Note that when calculating relative weights, the evidence source with the highest contribution will be fully retained, whereas the source with the lowest contribution will be disregarded, assigning it a weight of 0. This helps prevent decision-making from being influenced by incorrect information sources.

\begin{algorithm}
\caption{Weight calculation based on intrinsic fuzzy set}
\KwIn{the FOD $\Theta=\left\{A_1, A_2, A_3 \cdots A_k\right\}$, a set of objects to be identified $O=\left\{o_1, o_2, \ldots o_n\right\}$, a set of evidence sources $S=\left\{S_1, S_2, \ldots S_j\right\}$, BPAs of objects in object set $O_n$ from one source $S_j$}
\KwOut{ \( R_j \), \( W_j \)}
Initialize \( R_j \) to 0 \;
\For{each source $S_j \in\left\{S_1, S_2 \cdots S_j\right\}$}
{
    \For{each subject $o_i \in\left\{o_1, o_2 \cdots o_n\right\}$}
    {
        Take a BPA for an object $o_i$ from Source $S_j$ \;
        Compute the membership interval \( [\mu_A(x), 1 - v_A(x) ]\) according to Eq.(29-30) \;
        Calculate decision confidence \( d c_j \) according to Eq.(31) \;
        Calculate the correct decision contribution of the evidence source \(C_{jn}\) according to Eq.(32). \;
    }
    Update the reliability factor $R_j = R_j + r_{jn}$ \;
}
Determine the relative weight \( W_j \) using the reliability factor associated with each evidence source according to Eq.(34)\;
\Return $R_j$, $W_j$
\end{algorithm}

\section{Experiment and Analysis}
\label{section 4}
In recent years, pattern recognition has garnered significant attention due to its applications across various domains such as healthcare, finance, and robotics. This section aims to evaluate the reliability and superiority of the developed model in pattern recognition. The goal is to apply the model to various classification problems and assess its performance rigorously. By comparing its outcomes across different datasets and conditions, this study seeks to provide insights into its effectiveness and suitability for practical applications.
\subsection{Datasets}
This experiment incorporates five datasets: \textbf{Iris, Parkinsons, Connectionist Bench (CB), Fertility } and \textbf{Algerian Forest Fires (AFF)}, all sourced from the UC Irvine (UCI). The UC Irvine repository is renowned for its comprehensive collection of datasets, which span various domains and serve as standard benchmarks in the field of machine learning. Researchers worldwide rely on these datasets to develop, validate, and compare machine learning algorithms across diverse applications. The information regarding their categories, sample sizes, features, and missing values is provided in \hyperref[dataset table]{Table 1}.

\begin{table}[ht]
    \centering
    \caption{Summary of Experimental Datasets}
    \begin{tabular}{cccccc}
        \toprule
        & \textbf{Category} & \textbf{Sample Size} & \textbf{Features} & \textbf{Class} & \textbf{Subject Area} \\ \midrule
        1 & Iris & 150 & 4 & 3 & Biology \\
        2 & Fertility & 100 & 9 & 2 & Health and Medicine \\
        3 & Parkinsons & 197 & 22 & 2 & Health and Medicine \\
        4 & Connectionist Bench & 208 & 60 & 2 & Physics and Chemistry \\
        5 & Algerian Forest Fires & 244 & 14 & 2 & Biology \\
        \bottomrule
    \end{tabular}
    \label{dataset table}
\end{table}

\subsection{Comparative Models}
To further illustrate the accuracy of the FRI model in classification performance, a series of methods are used for comparison. These methods are generally classified into two main categories: (1) DST-related algorithms and (2) traditional machine learning algorithms.
\par
\subsubsection{Algorithms based on DS theory}
1. Murphy's method \cite{Murphy}: Murphy, in his study, proposed an idea of averaging the BPAs. By traversing every focal element within the existing evidence framework, he formulated a new evidence framework that encompasses all existing focal elements. The resulting BPA is an averaged structure. In Murphy's approach, the reliability of the evidence sources does not have a direct outcome, but it can be inferred by calculating the similarity between them and other evidence sources. Frequently occurring focal elements often result in a higher degree of BPA, reflecting this concept.
\par 2. Deng's method \cite{Deng}: Deng's research incorporates Murphy's averaging concept by also creating a new evidence framework that encompasses the known BPAs. However, Deng's approach to averaging BPAs is not an absolute average but is based on the similarity between evidence sources. Similarity among evidence sources is evaluated through the pignistic distance calculation, which, in theory, underscores the relevance of more dependable evidence sources.
\par 3. PCA \cite{PCA}: Principal Component Analysis (PCA) is a widely used technique for reducing dimensionality, aimed at converting high-dimensional datasets into lower dimensions while preserving the most critical features of the data. When applied to evidence analysis, PCA tends to retain evidence sources that capture the core characteristics of the evidence framework, while less relevant evidence sources, which do not align with the primary assessment focus, are given less weight.

\subsubsection{Classical Machine Learning Methods}

\par 1. SVM \cite{SVM}:
Support Vector Machines (SVMs) are a class of supervised learning algorithms distinguished by their principle of maximizing the margin between classes. SVMs identify a hyperplane within the feature space that serves to separate different categories, relying heavily on the support vectors, which are the data points closest to the decision boundary. Additionally, SVMs address non-linear classification issues through the use of kernel tricks, which map the data into a higher-dimensional space to achieve effective separation. This approach not only enhances the model's generalization capabilities but also circumvents the curse of dimensionality, making SVMs particularly effective for complex datasets.

\par 2. DT \cite{DT}: 
Decision Trees (DT) represent a machine learning technique that organizes decisions and their consequences into a tree structure. The method involves partitioning the data based on feature values to create subsets that are homogeneous with regard to the target variable. This tree-like model is straightforward to understand and is applicable for both classification and regression tasks.

\par 3. NaB \cite{NaB}:
Naive Bayes (NaB) is a probabilistic approach to classification that relies on the assumption of feature independence within each class. By evaluating the likelihood of each class based on the given features, it predicts the class with the highest probability. Despite the simplifying assumption of independence, Naive Bayes has proven to be highly effective across numerous practical scenarios.

\par 4. NMC \cite{NMC}: The Nearest Mean Classifier (NMC) is a simple algorithm that classifies a sample based on the proximity of its feature vector to the mean feature vectors of each class. It calculates the average feature vector for each class from the training data and assigns new samples to the class whose mean vector is closest. This method is easy to implement and works well when classes are distinctly separated in the feature space.

\par 5. KNN \cite{KNN}: The K-Nearest Neighbors (KNN) algorithm is a straightforward method for classification that assigns a sample to the category most common among its K nearest neighbors. By measuring distances from the sample to all points in the training set, KNN identifies the K closest points and classifies the sample according to the predominant class among them. Although KNN is easy to understand and works well with small to medium-sized datasets, it necessitates maintaining the complete training data for classification purposes.

\subsection{A Detailed Application Sample in Pattern Classification}
In this section, the proposed FRI method will be applied to the classification of the iris dataset, and its effectiveness will be explained in detail through comparisons with different algorithms.
\subsubsection{Dataset Description}
\begin{figure}
    \centering
    \includegraphics[width=1\linewidth]{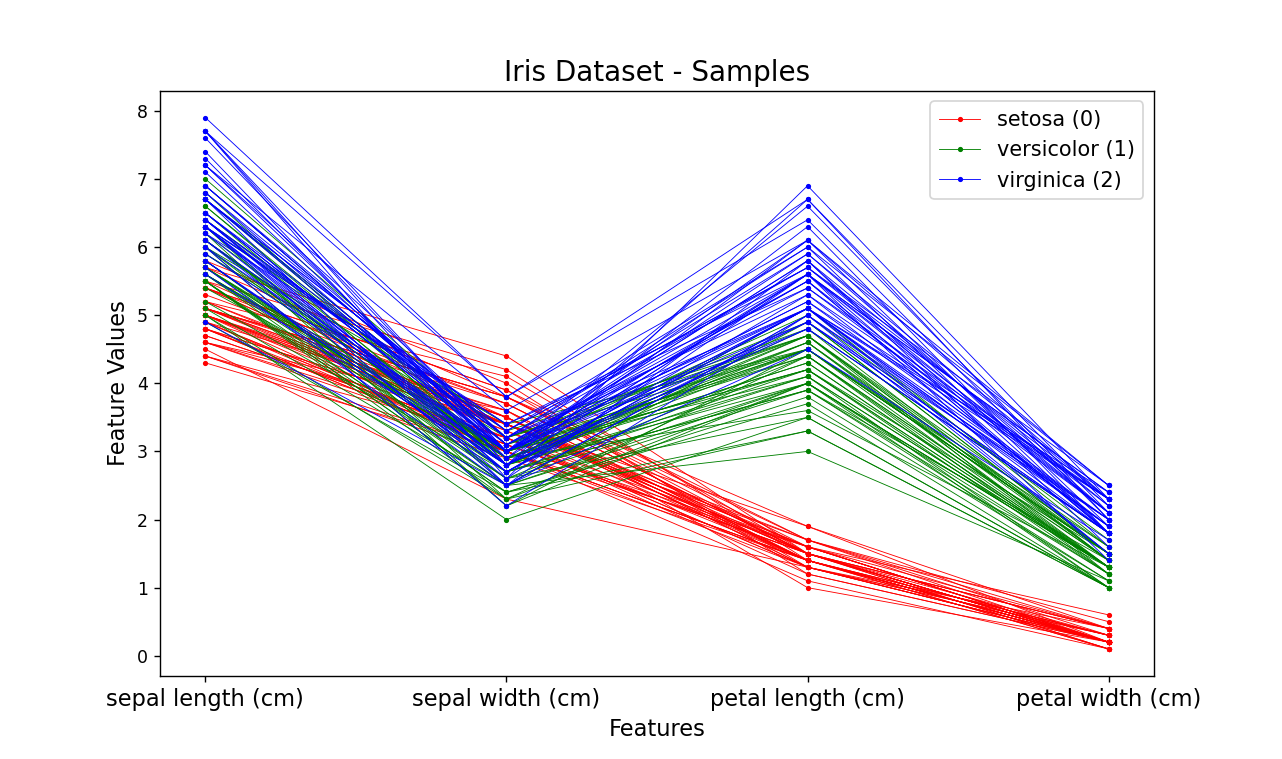}
    \caption{The values of four features of iris dataset samples}
    \label{iris}
\end{figure}

The Iris dataset from the UCI Machine Learning Repository is a well-known dataset in pattern recognition. It consists of 150 samples representing three species of iris flowers: Iris setosa (Se), Iris versicolor (Ve), and Iris virginica (Vi). Each sample includes four attributes: sepal length, sepal width, petal length, and petal width. Detailed information regarding these attributes for the different iris species can be found in \hyperref[iris]{Figure 3}.

\subsubsection{Implementation}
\noindent \textbf{Step 1:} 
Partition the Iris dataset such that 70\% of the data is used for training and 30\% is allocated for testing.\\
\noindent \textbf{Step 2:} Iterate through all training samples, generate BPAs through triangular fuzzy numbers, and then fuse them to obtain new BPAs. \\
\noindent \textbf{Step 3:} Calculate the reliability among different evidence sources according to the proposed method.\\
\noindent \textbf{Step 4:} Process the test set by applying reliability-based discounting to the evidence sources, then determine the classification accuracy after this processing step.

\subsubsection{Discussion}
\begin{figure}[h]
    \centering
    \begin{subfigure}[b]{0.25\linewidth}
        \centering
        \includegraphics[width=\linewidth]{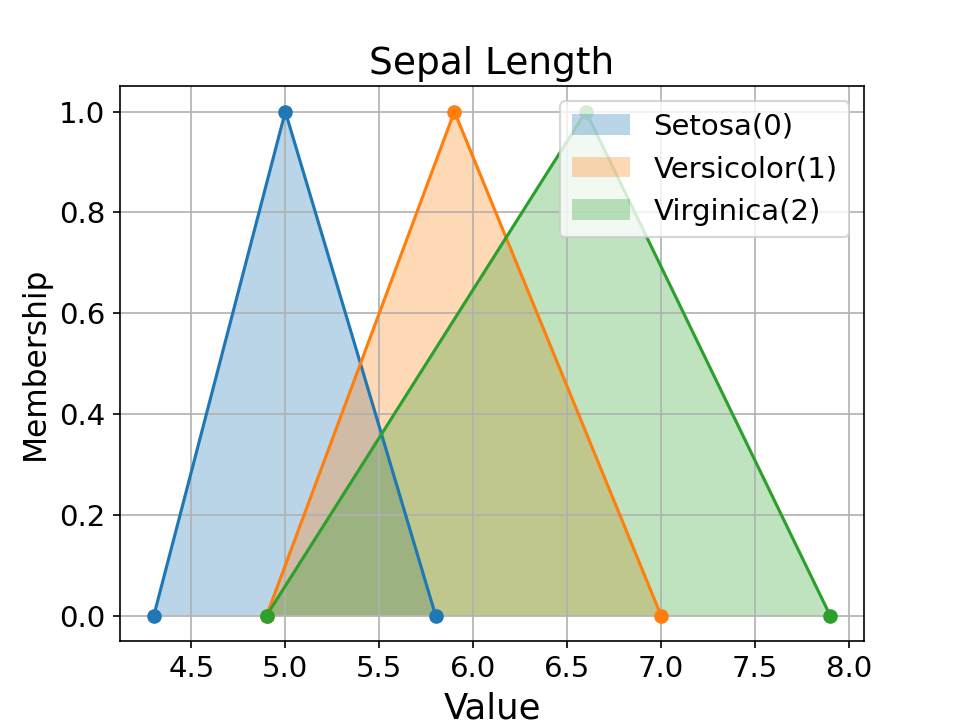}
        \caption{}
        \label{fig:feature1}
    \end{subfigure}\hfill
    \begin{subfigure}[b]{0.25\linewidth}
        \centering
        \includegraphics[width=\linewidth]{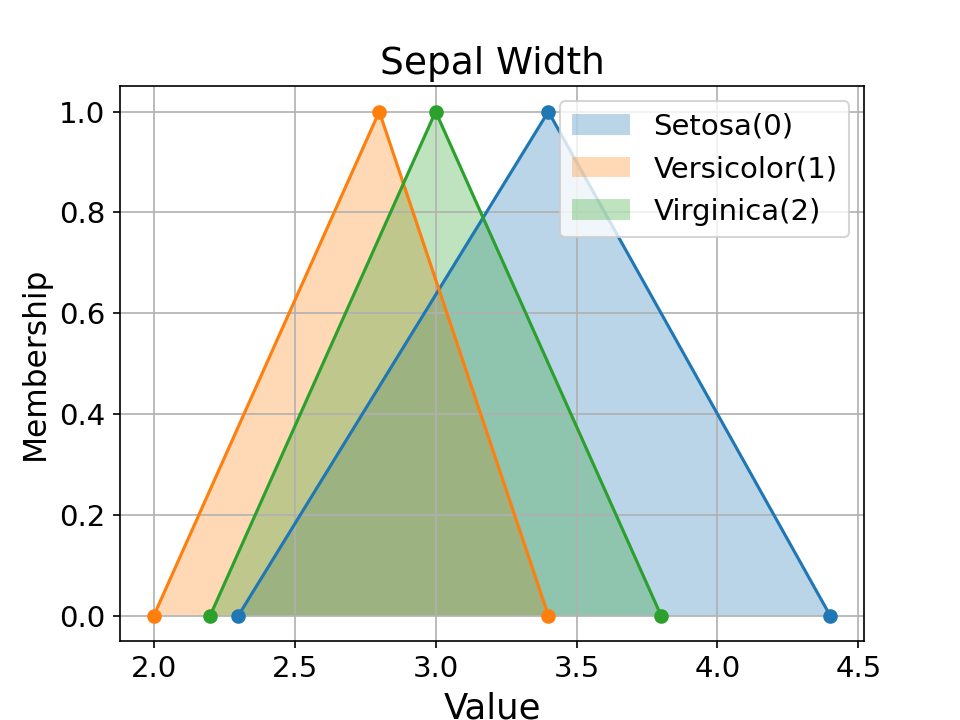}
        \caption{}
        \label{fig:feature2}
    \end{subfigure}\hfill
    \begin{subfigure}[b]{0.25\linewidth}
        \centering
        \includegraphics[width=\linewidth]{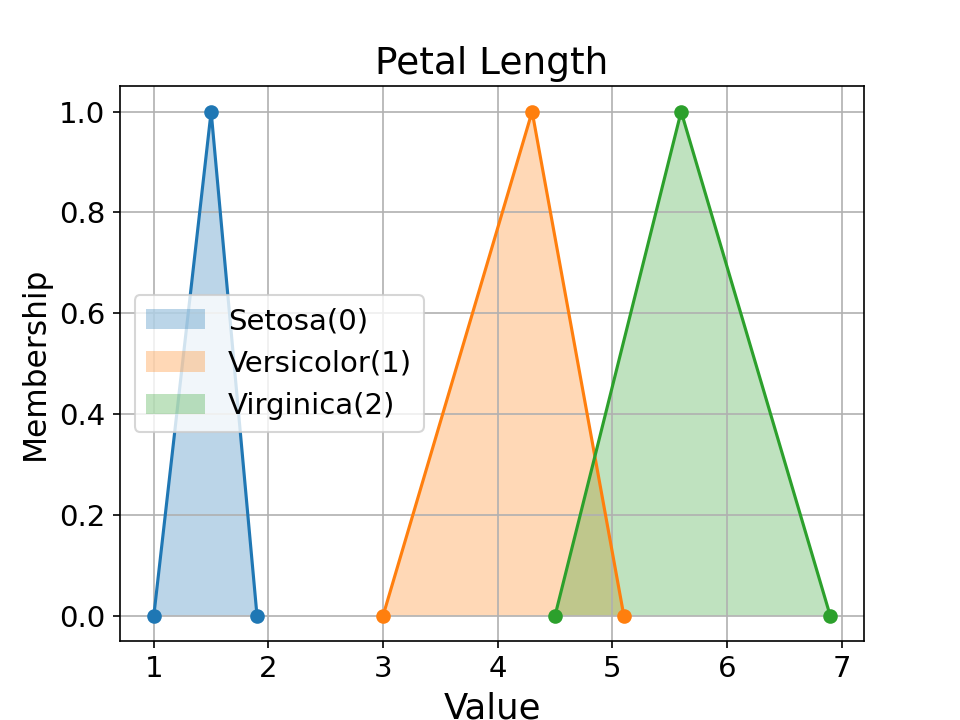}
        \caption{}
        \label{fig:feature3}
    \end{subfigure}\hfill
    \begin{subfigure}[b]{0.25\linewidth}
        \centering
        \includegraphics[width=\linewidth]{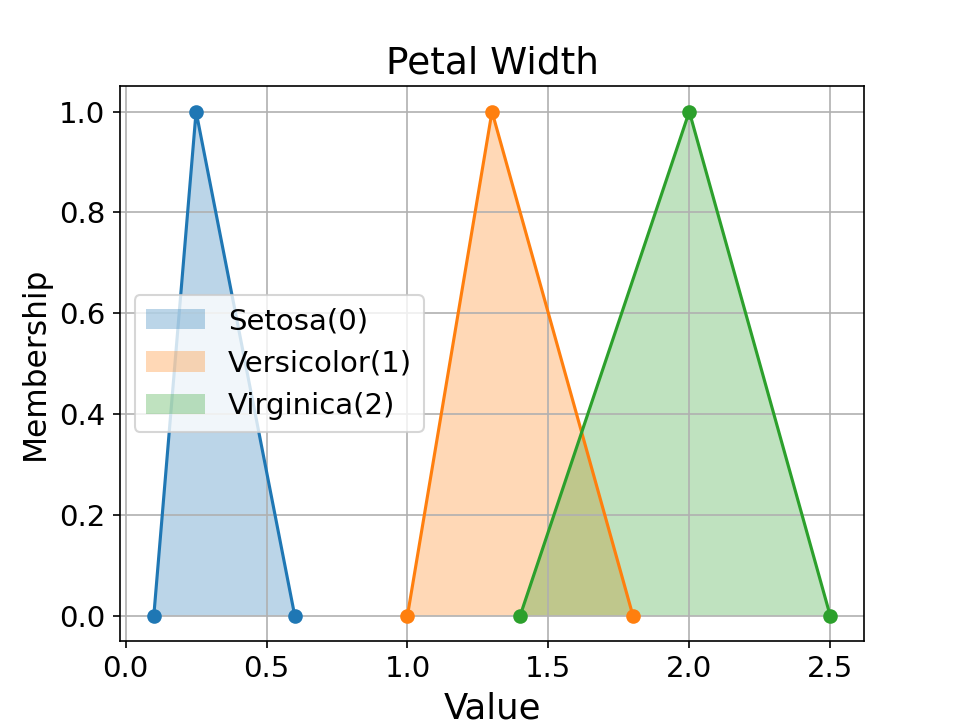}
        \caption{}
        \label{fig:feature4}
    \end{subfigure}
    \caption{Triangular fuzzy numbers of different features of iris.}
    \label{fig:overall}
\end{figure}

The maximum, average, and minimum values for different features of each Iris species are shown in \hyperref[fuzzy number]{Table 2}. Based on this, corresponding triangular fuzzy numbers and membership function graphs can be generated, as shown in \hyperref[fig:overall]{Figure 4}.

\begin{table}[h!]
\centering
\caption{Triangular Fuzzy Number of Iris}
\begin{tabular}{lcccc}
\toprule
        & Sepal Length & Sepal Width & Petal Length & Petal Width \\
\midrule
Setosa      &(4.3, 5.0, 5.8)&(2.3, 3,4 4,4)&(1.0, 1.5, 1,9)&(0.1, 0.25, 0.6)\\
Versicolor  &(4.9, 5.9, 7.0)&(2.0, 2.8, 3.4)&(3.0, 4.3, 5.1)&(1.0, 1.3, 1.8)\\
Virginica   &(4.9, 6.6, 7.9)&(2.2, 3.0, 3.8)&(4.5, 5.6, 6.9)&(1.4, 2.0, 2.5)\\
\bottomrule
\end{tabular}
\label{fuzzy number}
\end{table}

From Figure 2, it can be observed that the similarity between the three Iris species varies for different features. In terms of petal length (Fig. 2(c)) and petal width (Fig. 2(d)), the differences between the three Iris species are relatively pronounced, resulting in higher recognizability. Conversely, for sepal length (Fig. 2(a)) and sepal width (Fig. 2(b)), the similarity among the species is higher, which reduces their distinguishability.
\par During the training phase, since classical DST, Murphy, and Deng's methods do not directly define the reliability of evidence sources, the proposed FRI and PCA methods will be compared regarding evidence source reliability. Note that the reliability calculated by the proposed FRI is based on the contribution to the correct result, whereas PCA is derived solely from the similarity between evidence sources. The reliability of the features calculated by both methods is shown in \hyperref[reliability]{Table 3}. It can be observed that there are differences between PCA and FRI: for Petal Length (PL) and Petal Width (PW), since the differences among the three types of iris flowers are significant in these two features, they can serve as the main components for classification. Therefore, their contributions to the correct results are relatively high in both PCA and FRI. A notable difference lies in the treatment of Sepal Width (SW); FRI directly assigns it a reliability of 0, whereas PCA gives it a reliability of 0.91. This discrepancy stems from their respective calculation methods: FRI is primarily based on the comparison with the correct results, while PCA determines reliability solely based on the BPA itself, by assessing which feature best reflects the main characteristics. Additionally, in FRI, the least reliable evidence source might negatively impact the decision-making process, so assigning it a reliability of 0 is aimed at enhancing the overall classification reliability.

\begin{table}[ht]
    \centering
    \caption{Reliability of Features Calculated by FRI and PCA}
    \begin{tabular}{lcccc}
        \toprule
        & {Sepal Length} & {Sepal Width} & {Petal Length} & {Petal Width} \\ \midrule
        {FRI}  & 0.33 & 0.0 &  1.0& 0.90 \\
        {PCA}  &0.22 &0.91 &0.91 & 0.95\\
        \bottomrule
    \end{tabular}
    \label{reliability}
\end{table}

\par
After determining the reliability among different sources of evidence, a random iris sample from the training set was selected to test the distribution of BPAs across various methods. Its four feature values are [6.3, 3.3, 4.7, 1.6], and its true class is Versicolor. The initial BPAs generated for this sample after fuzzy triangular number processing, along with those processed under different methods, will be presented in \hyperref[iris BPAs]{Table 4}. \par Notice that because Murphy's and Deng's methods involve averaging the BPAs during the processing stage and then replacing the original BPAs with these averages, the processed four BPAs will become identical. After the fusion of each BPA, the PPT processing results for each method are shown in \hyperref[PPT BPAs]{Table 5}. The final results indicate that, without knowing the correct labels, the BPA value allocated to Ve (the correct label) by the FRI method is 0.9022, which is the highest. This demonstrates the reliability of FRI in making correct decisions.

\begin{table}[H]
\centering
\small 
\caption{ BPAs generated from the iris sample while it belongs to Sentosa}
\begin{tabular}{|c|p{1.5cm}|c|c|c|c|c|c|c|}
\hline
Method & Evidence & $\{S e\}$ & $\{V e\}$ & $\{V i\}$ & $\{S e, V e\}$ & $\{S e, V i\}$ & $\{V e, V i\}$ & $\{S e, V e, V i\}$ \\
\hline
\multirow{5}{*}{ DST }
& $m_1$ & 0.000 & 0.000 & 0.558 & 0.000 & 0.000 & 0.442 & 0.000 \\
& $m_2$ & 0.537 & 0.000 & 0.367 & 0.000 & 0.000 & 0.000 & 0.096 \\
& $m_3$ & 0.000 & 0.715 & 0.000 & 0.000 & 0.000 & 0.285 & 0.000 \\
& $m_4$ & 0.000 & 0.569 & 0.000 & 0.000 & 0.000 & 0.431 & 0.000 \\
& $m^*$ & 0.000 & 0.396 & 0.549 & 0.000 & 0.000 & 0.056 & 0.000 \\
\hline
\multirow{5}{*}{ PCA }
& $m_1$ & 0.000 & 0.000 & 0.125 & 0.000 & 0.000 & 0.099 & 0.776 \\
& $m_2$ & 0.491 & 0.000 & 0.336 & 0.000 & 0.000 & 0.000 & 0.173 \\
& $m_3$ & 0.000 & 0.652 & 0.000 & 0.000 & 0.000 & 0.260 & 0.087 \\
& $m_4$ & 0.000 & 0.540 & 0.000 & 0.000 & 0.000 & 0.409 & 0.052 \\
& $m^*$ & 0.008 & 0.605 & 0.272 & 0.000 & 0.000 & 0.112 & 0.003 \\
\hline
\multirow{5}{*}{ Murphy }
& $m_{1}$ & 0.134 & 0.321 & 0.231 & 0.000 & 0.000 & 0.290 & 0.024 \\
& $m_{2}$ & 0.134 & 0.321 & 0.231 & 0.000 & 0.000 & 0.290 & 0.024 \\
& $m_{3}$ & 0.134 & 0.321 & 0.231 & 0.000 & 0.000 & 0.290 & 0.024 \\
& $m_{4}$ & 0.134 & 0.321 & 0.231 & 0.000 & 0.000 & 0.290 & 0.024 \\
& $m^*$ & 0.003 & 0.632 & 0.325 & 0.000 & 0.000 & 0.040 & $1.385e^-06$ \\
\hline
\multirow{5}{*}{ Deng }
& $m_{1}$ &0.101 & 0.371 & 0.198 & 0.000 & 0.000 & 0.312 & 0.018   \\
& $m_{2}$ &0.101 & 0.371 & 0.198 & 0.000 & 0.000 & 0.312 & 0.018   \\
& $m_{3}$ &0.101 & 0.371 & 0.198 & 0.000 & 0.000 & 0.312 & 0.018   \\
& $m_{4}$ &0.101 & 0.371 & 0.198 & 0.000 & 0.000 & 0.312 & 0.018   \\
& $m^*$ & 0.001 & 0.746 & 0.214 & 0.000 & 0.000 & 0.039 & $3.427e^-07$ \\
\hline
\multirow{5}{*}{ FRI }
& $m_{1}$ &0.000& 0.000 & 0.182 & 0.000 & 0.000 & 0.142 & 0.673 \\
& $m_{2}$ & 0.000 & 0.000 & 0.000 & 0.000 & 0.000 & 0.000 & 1.000 \\
& $m_{3}$ & 0.000 & 0.715 & 0.000 & 0.000 & 0.000 & 0.285 & 0.000 \\
& $m_{4}$ & 0.000 & 0.511 & 0.000 & 0.000 & 0.000 & 0.387 & 0.102 \\
& $m^*$ & 0.000 & 0.835 & 0.030 & 0.000 & 0.000 & 0.135 & 0.000 \\
\hline
\end{tabular}
\label{iris BPAs}
\end{table}

\begin{table}[H]
\centering
\caption{BPAs after PPT operation}
\begin{tabularx}{\textwidth}{XXXX}
\toprule
Method & $\{S e\}$ & $\{V e\}$ & $\{V i\}$ \\
\midrule
DST & 0.0000 & 0.4237 & 0.5763 \\
PCA & 0.0091 & 0.6619 & 0.3290 \\
Murphy & 0.0026 & 0.6523 & 0.3451 \\
Deng & 0.0006 & 0.7656 & 0.2337 \\
FRI & 0.0000 & 0.9022 & 0.0978 \\
\bottomrule
\end{tabularx}
\label{PPT BPAs}
\end{table}

\subsection{Other Application Samples}
\par In this section, the generalizability of FRI across different datasets will be discussed. Since the previous section covered the specific processing and operations of FRI, this section will directly present results to compare FRI with other DS-based methods and some classical machine learning algorithms. This comparison aims to demonstrate the reliability and generalizability of FRI in classification tasks.

\par
\begin{figure}
    \centering
    \includegraphics[width=1\linewidth]{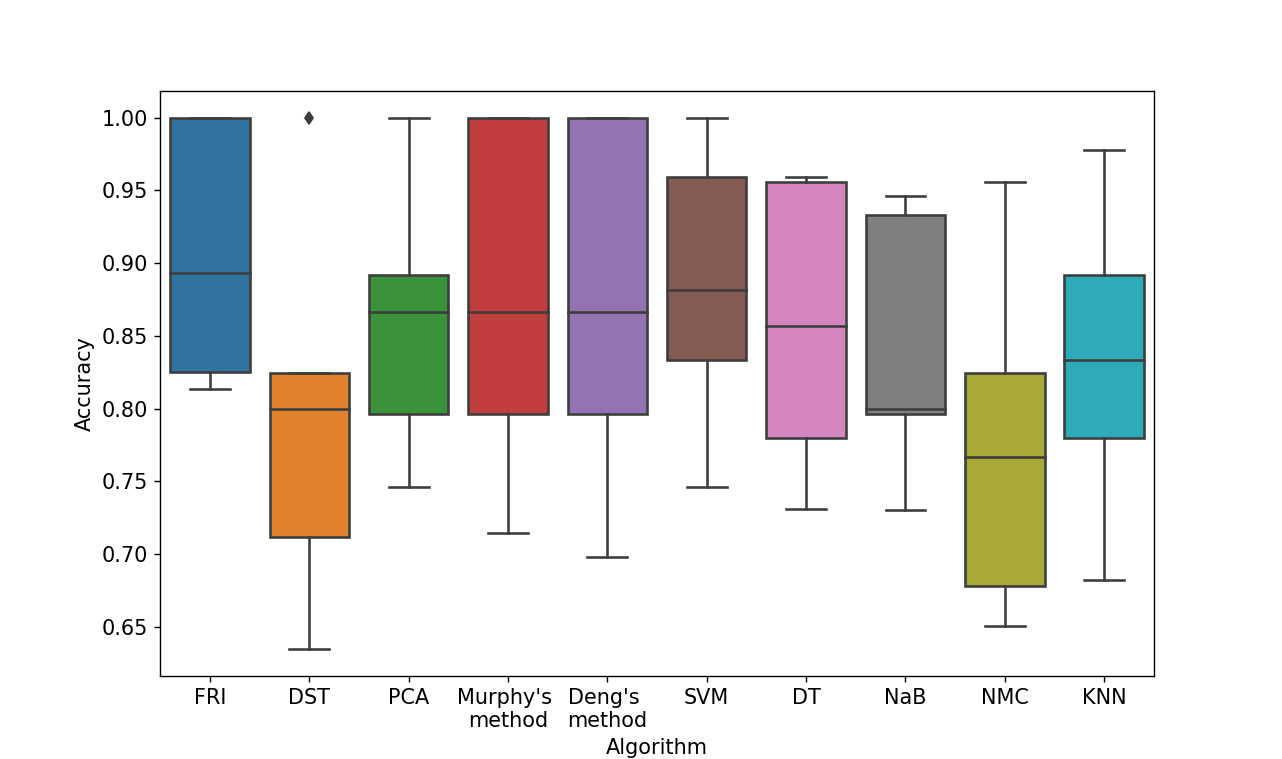}
    \caption{Comparison of classification accuracy between FRI and other algorithms on five data sets}
    \label{method results}
\end{figure}

 The datasets and algorithms used for this comparison have been introduced in Section 4.2. The performance of different algorithms on these datasets is shown in \hyperref[method results]{Figure 4}, with their specific accuracies detailed in \hyperref[accuracy]{Table 6}.

\par The box plot shows that the FRI algorithm generally performs the best, with a median accuracy close to 0.90 and a relatively concentrated accuracy distribution, indicating consistently high accuracy in most experiments. Additionally, the high lower bound of accuracy without significant outliers suggests good stability of the algorithm. The average accuracy in Table 6 further demonstrates the superiority of FRI. Murphy and Deng's methods exhibit outstanding upper limit performance on some datasets but have a broader distribution range. Among classical machine learning models, SVM displays the best stability.

\begin{table}[H]
\centering
\small 
\caption{Classification accuracies of different methods on various datasets}
\begin{tabularx}{\textwidth}{|l|X|X|X|X|X|X|}
\hline
 Datasets & Iris & Parkinsons & CB & Fertility & AFF & Average (\%) \\
\hline
\multicolumn{7}{|c|}{DS-based Algorithm} \\
\hline
FRI & 100.0 & 81.36 & 82.53 & 89.34 & 100.0 & 90.65 \\
\hline
PCA & 100.0 & 79.66 & 74.60 & 86.67 & 89.19 & 86.02 \\
\hline
DST & 100.0 & 71.19 & 63.49 & 80.00 & 82.43 & 79.42 \\
\hline
Deng & 100.0 & 79.66 & 69.84 & 86.67 & 100.0 & 87.23 \\
\hline
Murphy & 100.0 & 79.66 & 71.43 & 86.67 & 100.0 & 87.55 \\
\hline
\multicolumn{7}{|c|}{Classical ML Algorithm} \\
\hline
SVM & 100.0 & 88.13 & 74.60 & 83.34 & 95.95 & 88.40 \\
\hline
DT & 95.56 & 85.71 & 73.08 & 78.00 & 95.90 & 85.65 \\
\hline
NaB & 93.33 & 79.66 & 73.01 & 80.00 & 94.59 & 84.12 \\
\hline
NMC & 95.56 & 67.79 & 65.08 & 76.67 & 82.43 & 77.51 \\
\hline
KNN & 97.78 & 77.97 & 68.25 & 83.34 & 89.19 & 83.31 \\
\hline
\end{tabularx}
\label{accuracy}
\end{table}

\subsection{Discussion}
\par The superior stability and accuracy of the FRI can be attributed to the following factors:

1. The reliability of evidence sources in FRI is determined by a result-oriented approach rather than based on the similarity between evidence sources. This means that evidence sources that contribute positively to correct decisions are given higher reliability, while those with minimal or negative contributions are assigned lower reliability. This approach is more direct and efficient, eliminating the influence of highly similar but counterproductive evidence sources on correct decision-making.

2. The contribution based on intuitionistic fuzzy set is closely related to decision-making. This contribution not only comes from the BPA of correct decisions but is also related to the BPA of incorrect decisions. The interpretation of negative contributions makes this theory more comprehensive.

3. The connection with intuitionistic fuzzy set. The method mentioned in this experiment refers to the intuitionistic fuzzy set's definition in decision-making, which distinguishes it from other methods such as PPT. This provides a new perspective for decision-making in the DS theory and offers new ideas for future developments in DS-based decision-making processes.

\par Although the FRI algorithm proposed in this paper effectively addresses classification problems, its reliability calculation is based on the contribution to correct results. Therefore, FRI requires a supervised learning process during training, which means its accuracy is affected by the sample size. It does not involve the analysis of similarity between evidence sources, so its accuracy might decrease when the number of known samples is small. Additionally, FRI assigns a reliability of 0 to the evidence source with the smallest contribution, treating it as completely unknown. This makes FRI particularly suitable for recognition problems with a large number of features. When the features include those that negatively impact the correct result, FRI can demonstrate better performance.

\section{Conclusion}
\label{conclusion}
This paper introduces a novel method for assessing the reliability of evidence sources, termed the Fuzzy Reliability Index (FRI). The FRI approach leverages intuitionistic fuzzy set theory. Initially, a set of transformation rules is established by exploring the similarities between IFS and DST. Subsequently, the method defines and quantifies the contribution to accurate decision-making, considering both positive and negative impacts. These contributions affect the assessed reliability of evidence sources. In cases where negative contributions are present, the reliability is calculated based on the relative magnitude of these contributions, with normalization adjusted according to the size of the difference intervals.
\par In the experimental section, several information fusion algorithms based on DS theory and some classic machine learning algorithms were compared with FRI to demonstrate its reliability and superiority in pattern classification problems. The experiment first conducted a detailed analysis using the iris dataset, including the generation of BPAs, determination of reliability, discounting processing, and final result analysis. To verify the generalizability of FRI, additional datasets from different fields were used for a comprehensive comparison. The results ultimately demonstrated that the proposed FRI has advantages across various datasets and is more effective than other known methods.
\par The primary contribution of this paper is the development of a decision quantification rule and a contribution measurement method rooted in intuitionistic fuzzy sets, distinguishing it from the PPT algorithm. The proposed FRI fusion algorithm is particularly suited for supervised learning with large datasets. It demonstrates exceptional performance in cases where there are many features and high overlap in feature values among samples, making differentiation challenging. In the future, FRI will focus on considering the similarity between different evidence sources, rather than solely being result-oriented. Additionally, more reasonable standards will be established in the calculation of reliability from contributions to enhance the applicability of the FRI fusion algorithm.

\section*{CRediT authorship contribution statement }
\textbf{Juntao Xu}: Conceptualization, Data curation, Formal analysis, Methodology, Resources, Software, Validation, Visualization, Writing – original draft, Writing – review \& editing. \textbf{Tianxiang Zhan}: Conceptualization, Methodology, Resources. \textbf{Yong Deng}: Funding acquisition, Investigation, Project administration, Resources, Supervision, Writing – review \& editing.

\section*{Declaration of competing interest }
The authors declare that they have no known competing financial interests or personal relationships that could have appeared to influence the work reported in this paper.

\section*{Data availability}
No data was used for the research described in the article.

\section*{Acknowledgments}
The work is partially supported by National Natural Science Foundation of China (Grant No. 62373078).

\bibliographystyle{elsarticle-num}
\bibliography{ref}

\end{document}